\documentclass[10pt, a4paper]{article}
\usepackage{authblk}
\usepackage{lrec-coling2024} 
\usepackage{xcolor}
\usepackage{caption}

\title{BAN-PL: a Polish Dataset of Banned Harmful and Offensive Content from Wykop.pl web service\\
\vspace{10pt}
\small\textcolor{orange}{WARNING: This paper contains examples of language and content that may be offensive, discriminatory, or hateful in nature.}}

\name{Anna Kołos$^{\ddag,\ast}$, Inez Okulska$^{\ddag,\ast}$, Kinga Głąbińska$^{\ast}$, Agnieszka Karlińska$^{\ast}$, \\
    \bf{\large Emilia Wiśnios$^{\ast}$}, \bf{\large Paweł Ellerik$^{\dagger}$}, \bf{\large Andrzej Prałat$^{\dagger}$}} 
\address{$^{\ddag}$equal contribution \\ $^{\ast}$NASK National Research Institute (Warsaw), $^{\dagger}$Wykop.pl}

\abstract{
Since the Internet is flooded with hate, it is one
of the main tasks for NLP experts to master automated online content moderation. However, advancements in this field require improved access to publicly available accurate and non-synthetic datasets of social media content. For the Polish language, such resources are very limited. In this paper, we address this gap by presenting a new open dataset of offensive social media content for the Polish language. The dataset comprises content from Wykop.pl, a popular online service often referred to as the \textit{Polish Reddit}, reported by users and banned in the internal moderation process. It contains a total of 691,662 posts and comments, evenly divided into two categories: \textit{harmful} and \textit{neutral} (\textit{non-harmful}). The anonymized subset of the BAN-PL dataset consisting on 24,000 pieces (12,000 for each class), along with preprocessing scripts have been made publicly available. Furthermore the paper offers valuable insights into real-life content moderation processes and delves into an analysis of linguistic features and content characteristics of the dataset. Moreover, a comprehensive anonymization procedure has been meticulously described and applied. The prevalent biases encountered in similar datasets, including post-moderation and pre-selection biases, are also discussed.  \\ \newline 
}

\begin{document}

\maketitleabstract

\section{Introduction}


Given the pervasive nature of offensive language on the Internet, proficiently managing automated online content moderation stands as a paramount objective for NLP experts. However, the main challenge remains the accessibility of accurate datasets, a task complicated by the inability to synthetically represent them. The wide spectrum of linguistic inventiveness employed by cyberbullies, coupled with fractured syntax, necessitates at least a large language model to generate linguistically convincing offensive samples. Yet exactly these models block such content, making it hardly possible to analyze or create it (\citet{zhou}). Hence, training data should be gathered from the everyday flood of new content posted online. However, finding a reliable and representative source is still a major issue, especially for low-resource languages. 

While there has been significant growth in the availability of resources in recent years, such as the inclusion of 25 languages in the Catalog of Abusive Language Data\footnote{\url{www.hatespeechdata.com}} (\citet{vidgen2020}), the development of resources for offensive content has been uneven, with only one entry listed for the Polish language.

In this paper, we introduce BAN-PL, a Polish language dataset of banned offensive content, to address this gap in available resources. 
Notably, it is one of the few publicly available datasets (\citet{ljubesic-etal-2018-datasets}) that includes content reported by community members (here users of the service Wykop.pl), identified as offensive, and subsequently deleted by professional content moderators.

Wykop.pl was established in December 2005 as an equivalent to the digg.com web service (\citet{digg-wiki}). Often referred to as the \textit{Polish Reddit} (\citet{Wojcik-2021}), it is currently listed as one of the top 10 most popular social networking services (SNS) in Poland, boasting approximately 3 million real users as of February 2023. These users generate around 30,000 new posts and comments daily (\citet{wirtualnemedia-2023}). The content is under constant monitoring by trained moderators from the user community. Additionally, every user can independently flag any piece of content within an internal taxonomy of over 20 different ban reasons.\footnote{\url{www.wykop.pl/standardy-moderacji}} If content is flagged by users, it must undergo review by a moderator to either be approved as \textit{non-harmful} or removed from the website.

Besides introducing a new dataset and analyzing its most critical linguistic features and characteristics, the paper also provides valuable insights into real-life content moderation processes. Furthermore, a comprehensive anonymization procedure, exceeding the typical masking of usernames and URLs, has been meticulously described and applied. The prevalent biases encountered in similar datasets, including post-moderation and pre-selection biases (see \ref{discussion}), are also discussed.


\section{Related work}

The field of automated offensive language detection has seen significant efforts in the NLP community (for an overview of the latest developments in this area, see \citet{jahan2023, alrashidi2022}). A crucial factor for further progress is a widespread access to diverse, publicly available datasets of social media content. To detect offensive language in English, researchers commonly use datasets from platforms like Twitter (\citet{waseem2016hateful, Davidson, Founta}), as well as Wikipedia talk pages (\citet{Wulczyn}). Some papers also focus on hateful comments on platforms like Facebook and YouTube (\citet{Hammer, Salminen}). \citet{Zampieri2019} created a hierarchical dataset from Twitter called the Offensive Language Identification Dataset (OLID), which was adopted for SemEval-2019 Task 6, aimed at identifying and categorizing offensive language in social media (\citet{semeval2019}). Additionally, a dataset of content from banned communities on Reddit (RAL-E) has been employed to train the BERT large language model for detecting abusive language, leading to the creation of HateBERT (\citet{caselli2020}).

The diversity of existing datasets encompasses various aspects, including different data sources with: (i) varied labels and modes of annotation, such as hate speech, offensiveness, aggression, racism, sexism, and toxicity; (ii) distinct targets of attacks, including personal attacks, attacks on women, and attacks on migrants; (iii) differing numbers of classes, ranging from binary to multi-label classification; (iv) varying degrees of class balance, where, in many cases, the neutral class (representing non-harmful content) significantly outweighs the harmful class or classes (see hate-related dataset characteristics in \citet{MacAvaney-2019}). A specific case involves datasets of content flagged for review by online communities and subsequently removed by trained moderators. There are very few publicly accessible datasets like this. An example is Slovenia's MMC and Croatia's STY, which encompass news comments, including content removed during the moderation process (\citet{ljubesic-etal-2018-datasets}). Both datasets are publicly available, albeit in encrypted form. A small portion of the content in these datasets, approximately 8\% in MMC and around 2\% in STY, has been deleted by moderators. For a comprehensive overview of existing corpora up to 2021, see (\citet{Poletto2021}).

Until recently, only two datasets containing offensive social media content, including hate speech, were publicly available for the Polish language. The first dataset was acquired in 2017 (\citet{troszynski}), but it was made accessible on HuggingFace only in 2021.\footnote{\url{https://huggingface.co/datasets/hate_speech_pl}} The second one was originally introduced in task 6 of the PolEval2019 competition.\footnote{\url{https://2019.poleval.pl/index.php/tasks/task6}} It is part of the KLEJ Comprehensive Benchmark for Polish Language Understanding, the equivalent of the well-known GLUE benchmark, which encompasses 9 general natural language understanding tasks including cyberbullying detection (CBD) (\citet{klej}). Detailed information regarding data collection and preprocessing procedures can be found in~\citet{poleval_ptaszynski}, while enhancements to the dataset have been elaborated upon in \citet{ptaszynski2023}.

State-of-the-art results in the CBD KLEJ task (F1 score = 76.1) were achieved by the TrelBERT model pre-trained on Twitter data (\citet{trelbert-2023}). The authors of the model created an additional test dataset called "harmful\_tweets\_1k"\footnote{\url{https://github.com/deepsense-ai/trelbert}} to assess the generalisation of their solution to a broader distribution of Twitter data. 
The results indicate that TrelBERT, when used in a real-world scenario, has a higher ability to detect harmful content than larger models trained on general-domain corpora, e.g., HerBERT or Polish RoBERTa.\footnote{\url{https://klejbenchmark.com/leaderboard/}}

Among the resources available for the Polish language, it is also worth mentioning HateSpeech,\footnote{\url{https://ws.clarin-pl.eu/hatespeech}} a tool for personalized hate speech recognition created by CLARIN-PL.\footnote{\url{https://clarin-pl.eu/index.php/en/home/}} It was trained on datasets obtained as part of the Wikipedia Detox project\footnote{\url{https://meta.wikimedia.org/wiki/Research:Detox/Data_Release}} and allows for a broader analysis of offensive language (\citet{KOCON2021102643, KAZIENKO202343}).


\section{Data source}

In a joint project between the institution represented by the authors and the web service Wykop.pl, we were able to collect a unique dataset of harmful online content consisting of texts banned and labeled by professional moderators. The BAN-PL dataset comprises a total of 691,662 pieces of content, of which 345,831 were assigned to the "harmful" class and the exact same number to the "neutral" class. Non-textual material has been removed. The "harmful" class comprises posts and comments banned by moderators during the period spanning from January 2019 to April 2023. The collection of "neutral" data was executed between March and August 2021, June and November 2022, and January and June 2023. 

\subsection{Moderation scheme}
\label{subsection:moderation}

The moderators have been recruited from the Wykop.pl users community and have undergone dedicated training. Currently, there are 14 of them. Each piece of content, flagged by the users, is independently assessed by five distinct moderators, and the final score is based on the majority voting. According to information provided by Wykop.pl, the 3:2 vote split, representing an indecision, is rare. In most cases, a consensus decision is reached, with the majority aligning with a particular rating, as represented by a 5:0 or a 4:1 vote split.

In addition, the platform allows users to appeal decisions made by the moderation team. This results in a reassessment of a moderation rating. However, the content in question is then automatically assigned only to moderators who were not involved in the initial assessment of the reported post or comment. They have the authority to reject the appeal or accept it by reinstating the removed content to the site. Further appeals will be escalated to the service administrators, removing them from the direct jurisdiction of the moderation team.

In overall, approximately 2\% to 5\% of the platform's content is reported to the moderation team, which consists of 30,000 to 60,000 submissions monthly. Of the content reported by users, 49\% adhere to website policies. Roughly 15\% are related to spam and flooding, 10\% involve inappropriate content, and 7\% pertain to incorrect tags. A small fraction (1\% to 2\%) of submissions is re-evaluated, with 3\% to 5\% deemed as valid appeals.

\begin{figure}[!h]
\centering
\includegraphics[width=7.5cm]{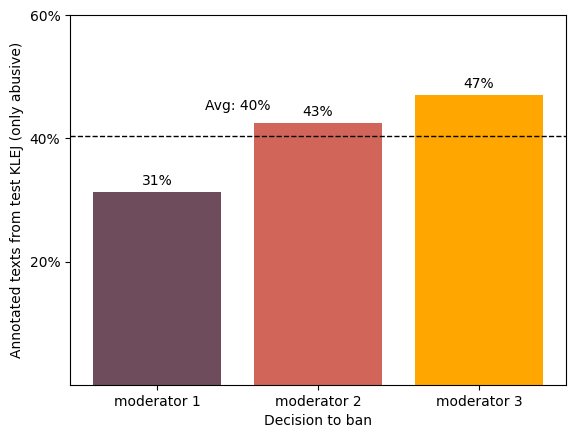}
\caption{Percentage of samples from the KLEJ test set labeled as offensive by three Wykop.pl moderators}
\label{ban_not_ban_ann}
\end{figure}

Since we are unable to evaluate the quality of the moderators' work according to the guidelines applied to the evaluation of manually annotated datasets, we asked three moderators to re-annotate 134 samples of the "harmful" class from the KLEJ test dataset. The results of this experiment showed that only 40\% of the tweets were labelled as offensive, with varying percentages among the annotators. (Fig.~\ref{ban_not_ban_ann}).

The inter-annotator agreement yielded an avarage score of 0.59 based on Fleiss' and Randolph's kappa multi-rater agreement measure. Given that the annotators were equally trained by Wykop.pl and followed the guidelines of the website's policy, a higher score could have been expected. However, in their everyday moderation process, the moderators have access to the context which was not granted in this case (the isolated text samples were sent via a Google form for scoring). The agreement scores did not differ significantly between pairs of annotators yielding in average 0.59 with 2.64 STD. This observation suggests a notable disparity between Wykop's internal moderation policy and the guidelines commonly employed by professional annotators for labeling datasets, as exemplified within the CBD task as part of the KLEJ benchmark.

\subsection{Data collection process}

For the purpose of offensive language detection, we included in the dataset both content banned for inciting hatred or violence (i.e., "Propagation of hatred or violence, drastic content") and content including personal attacks (i.e., "Attacks me or violates my personal rights," "Attacks me," "Attacks others"). In order to streamline the analysis and classification process, we have merged these categories into a single class labeled as "harmful." This consolidated class includes the following types of posts and comments that violate the platform's policy:

\begin{enumerate}
\itemsep-0.1em
\item Content inciting hatred or violence towards others, as well as attacking individuals/groups based on protected characteristics such as gender, race, ethnic origin, nationality, religion/belief/worldview, disability, age, sexual orientation.
\item Content promoting violence, graphic material, or advocating the spread of hatred, including racism, xenophobia, national conflicts, or homophobia.
\item Graphic or literal depictions of violence, such as torture, stoning, mistreatment of humans or animals, murder, suicide, or expressions of hatred.
\item Materials that damage one's reputation, including false information, vulgar insults, defamation, verbal abuse, intimidation, blackmail, and threats.
\item Personal attacks and targeting individuals based on protected characteristics, such as gender, race, ethnic origin, nationality, religion/beliefs/worldview, disability, age, or sexual orientation.
\end{enumerate}

In addition to the aforementioned categories, we have also included content marked as "inappropriate" in our dataset. This category is broad and encompasses various types of content that are deemed unsuitable, including incitement to commit a crime or suicide, pornographic materials, content promoting violence, promotion of drugs, etc. Therefore, a supplementary task of classification of the "inappropriate" content has been performed (see details in section \ref{Classification}). 


This initial "harmful" class consisted of 379,511 pieces of content. The content banned for inciting hatred or violence (n = 75,396) comes from 20,130 unique users. The content that was banned due to containing personal attacks (n = 72,990) originates from 19,076 unique users. The content marked as "inappropriate" (n = 231,125) came from 37,596 unique users. To obtain a balanced dataset for the binary classification task, the contrasting neutral class was gathered from the main page of the web service. Only entries and comments published at least 48 hours prior to scraping were considered, taking into account the moderation dynamics of the platform. The main page, being highly visible to the public, typically ensures that any harmful content is reported by the user within first 24 hours. Therefore, samples that have not been banned or reported within 48 hours can be regarded as "neutral". The initial "neutral" class consisted of 560,750 unique content pieces, covering various topics.

\subsection{Classification}
\label{Classification}

Content flagged as "inappropriate" by moderators largely overlaps with content that includes personal attacks and incitement to hatred or violence. However, this category was less specific and included a broader spectrum of phenomena. To maintain data consistency, we decided to perform an automated classification of content flagged as inappropriate. We used content banned for inciting hatred or violence, content banned for personal attacks (n = 148,386), and a random sample of "neutral" content (n = 148,386) as training data. The choice of the model was preceded by experiments using the large Polish model RoBERTa (\citet{polish_roberta}) and TrelBERT. The best model turned out to be the one using the average of the last hidden state from RoBERTa large as the pooled output of the model. This output was activated with ReLU and went to a fully connected layer. The loss function used was BCEWITHLOGITSLOSS (combination of sigmoid layer and BCELoss function). The learning rate used in the experiment was 1.5e-6. The model was fine tuned on the "harmful" class and the neutral samples. The dataset was split into three subsets, with 75\% of the data allocated for training, 12.5\% for validation, and a further 12.5\% for testing purposes. When applied to the flagged "inappropriate" content, the model classified 85.43\% of the observations as "harmful" and 14.57\% as "non-harmful" (F1 score = 0.92). We, therefore, only included the observations classified as harmful in the dataset. 


The final harmful class includes 148,386 posts or comments flagged by moderators as inciting violence or personal attacks, and 197,445 posts or comments flagged as "inappropriate" and classified as "harmful," for a total of 345,831 pieces of content. The small portion of data (33,680 posts or comments) from the class initially flagged as "inappropriate" but classified as "not harmful" was excluded from the dataset.
In order to maintain balance between harmful and neutral content within the dataset, a random selection of neutral post and comment was made to match the number of harmful pieces of content (n = 345,831). 

\section{Data anonymization}
\label{section:anonymization}

Due to the presence of personal data, such as home and email addresses, phone numbers or PESEL numbers (identification number assigned to individuals in Poland and used for administrative purposes), within the dataset, a comprehensive anonymization approach was imperative. Usernames and hyperlinks have also been masked. In order to prevent the further spread of offensive content, the anonymization process included the surnames and pseudonyms of individuals targeted by such content. These were primarily public figures, politicians, authors, and celebrities.


Achieving complete anonymization, encompassing the elimination of all contextual information linked to these individuals that could potentially facilitate their identification, proves unattainable without resorting to substantial reductions, subsequently engendering distortion in the analyzed data. 

In light of this, we have opted to conceal all surnames and pseudonyms (particularly those of social media influencers, prominent YouTubers, and streamers) through the implementation of placeholders, denoted as either [surname], or [pseudonym]. Usernames mentioned in the main body of the text, unaccompanied by the customary "@" symbol, have been replaced with the pseudonym tag as well. In order to maintain contextual integrity, as well as the usability and relevance of the data for further research purposes (especially from a sociolinguistic perspective), we decided to retain first names, indicating whether the target is male or female and frequently bearing eponymical significance, names of public organizations and political parties. In agreement with Wykop.pl data providers, it has been decided that fictional characters, historical figures and deceased individuals are to remain non-anonymized. 

Notably, there is no readily available solution for the Polish language that would provide a high level of precision in such complex anonymization procedures with respect to social media content.  Thus, we employed a medley of techniques and tested various approaches. Our anonymization pipeline encompasses a robust combination of Named Entity Recognition-, dictionary- and rule-based approaches. It is based on the PrivMasker tool\footnote{\url{https://github.com/ZILiAT-NASK/PrivMasker}} and PolDeepNer2.\footnote{\url{https://github.com/CLARIN-PL/PolDeepNer2}} PrivMasker, built upon the spaCy library (\citet{spacy2}), enables automatic detection and masking of selected categories of personal data. PolDeepNer2, a neural model based on RoBERTa language model, is one of the latest and most effective (it scored 0.899 F1-score measure on the PolEval 2018 set) NER tools for the Polish language (\citet{poldeepner2018, poldeepner2021}). Additionally, hyperlinks and mentions, i.e. usernames preceded by the “@” symbol, were masked using regex formula. 

Nonetheless, these tools exhibited limitations in their scope, as they failed to recognize pseudonyms and differentiate between authentic and fictitious or historical characters. 
Instances where names had been intentionally or inadvertently altered by users posed further challenges for the models. A process of fine-tuning became essential to achieve the desired results. Consequently, as an initial measure, we opted to release a subset of 24,000 posts and comments, each of which underwent meticulous manual verification of its anonymization process. Each piece of content was evaluated by one annotator and verified by a super-annotator according to detailed guidelines. In total, seven annotators and two super-annotators (linguists with experience in annotating social media data) participated in this process. Simultaneously, based on a review of prevalent surnames and hashtags across the entire dataset and coupled with expert knowledge, dictionaries of historical and fictional figures, as well as popular pseudonyms were developed. 

Sensitive information was replaced by the following tags: username, URL, surname (including user-modified ones and adjectives derived from surnames), pseudonym, address (including street names, house numbers, postal codes), email, phonenumber, number (other numbers, e.g., PESEL). Examples of their use are given in Tab.~\ref{table:examples_anon}.

\begin{table*}
\caption{Examples of anonymized content of the BAN-PL dataset*}
\label{table:examples_anon} 
\centering
\scalebox{0.8}{
\begin{tabular}{|p{9.5cm}|p{9.5cm}|}
\hline

     \textbf{Original sample after anonymization (in Polish)}	&\textbf{English translation} \\ \hline
     
    będziesz pan wisiał, panie [surname] & you will be hanging, mr [surname] \\\hline
    
    [surname] to glupia c**a i nie nalezy sie jej ten nobel & [surname] is a stupid c**t and she doesn’t deserve this nobel prize\\\hline
    
    \{USERNAME\}: pedała ci żal że się nad ty[m] [pseudonym] użalasz? & \{USERNAME\}: do you pitty a f****t since you feel sorry for this [pseudonym]?\\\hline

    \#[pseudonym] k***o pedophilu. Nie jesteś streamerem. Nie liczysz się w świecie YouTube. Płace 5.000 zł za adres tej k***y \#patostreamy & \#[pseudonym] you wh**e pedophile. You’re no streamer. You don't count in the YouTube world. I’m paying 5,000 PLN for this wh***’s address \#patostreamy \\\hline
    
    Do tego świra-pedalarza w lateksach z [address] w Poznaniu - wyk*****j cwelu i sam patrz do tyłu. Sorry za zj***nie średniej idioto. \#rower \#pedalarze & To this kook-pedal pusher in latexes from [address] in Poznan - f**k off w****r and look back yourself. Sorry to f*** up your average idiot. \#bike \#pedalpushers \\\hline
    
    \{URL\} c**j Wam na stale, śmiecie z jutuba, czyli urywek ostatniego streamu! \#[pseudonym] \#[pseudonym] \#patostreamy \#[pseudonym]  \#patostreamy & \{URL\} f**k you forever, jutube trash, that is a snippet from the last stream! \#[pseudonym] \#[pseudonym] \#patostreamy \#[pseudonym] \#patostreamy \\\hline
    
    [phonenumber] napalona agatka lat 16 lubi BDSM & [phonenumber] horny agatka aged 16 likes BDSM
\\\hline
\end{tabular}
}
\vspace{0.1cm}
\caption*{\textsuperscript{*}This table contains examples of hate speech and vulgar personal attacks. The authors do not support the use of harmful language, nor any of the harmful representations quoted above.}
\end{table*}

\section{Data specificity}

\subsection{Wykop.pl web service}
\label{section:Wykop}

As an online platform for news aggregation, Wykop.pl is specifically designed to cater to the interests of the internet audience, offering a wide range of content spanning various topics such as sports, economy, trending political affairs, and viral Internet phenomena. The primary feature of Wykop.pl is its user-generated content, which allows users to submit and share news stories, articles, and other forms of media. This user-driven approach ensures that the platform remains dynamic and up-to-date, reflecting the diverse interests and perspectives of its user base. 

According to data from verified accounts, the user base of Wykop.pl predominantly falls within the 18-45 age range, with a notable contribution from individuals aged 15-24 (\citet{Wirtualnemedia-2014}). Furthermore, the platform exhibits a higher proportion of male users. What is more, Wykop.pl represents a relatively more closed and potentially more homogeneous community when compared to the broader population of online users who generate content on other platforms. This observation is supported by the development of its unique sociolect and semiotic sign system, indicating a distinctive linguistic and communicative environment within the platform (\citet{rak, Sowinski_2018}). 

One distinctive aspect of Wykop.pl is its voting system, which enables users to participate in the curation of content. Users have the ability to vote on submitted content, either by "digging" or "burying" it. By digging a post, users express their approval and contribute to its visibility, while burying a post indicates disapproval and reduces its prominence within the platform. 

Another distinctive feature is the folksonomy, a collaborative tagging system system, which enhances content organization and discoverability. 
Tags serve as metadata, providing additional context and enabling easier navigation and search on the platform. Furthermore, the Wykop.pl community leverages tags to provide context, particularly when it comes to humor and sarcasm. Users on the platform often employ tags to indicate the intention behind their posts, allowing others to interpret the content correctly based on their collective experience and understanding (\citet{Sowinski_2018}). Unlike platforms like Twitter, where tags are often integrated within the utterances themselves due to character limits, on Wykop.pl, the tags typically appear at the end of the utterances.

\subsection{Dataset in numbers}

\begin{figure}[!h]
\centering
\includegraphics[width=8cm]{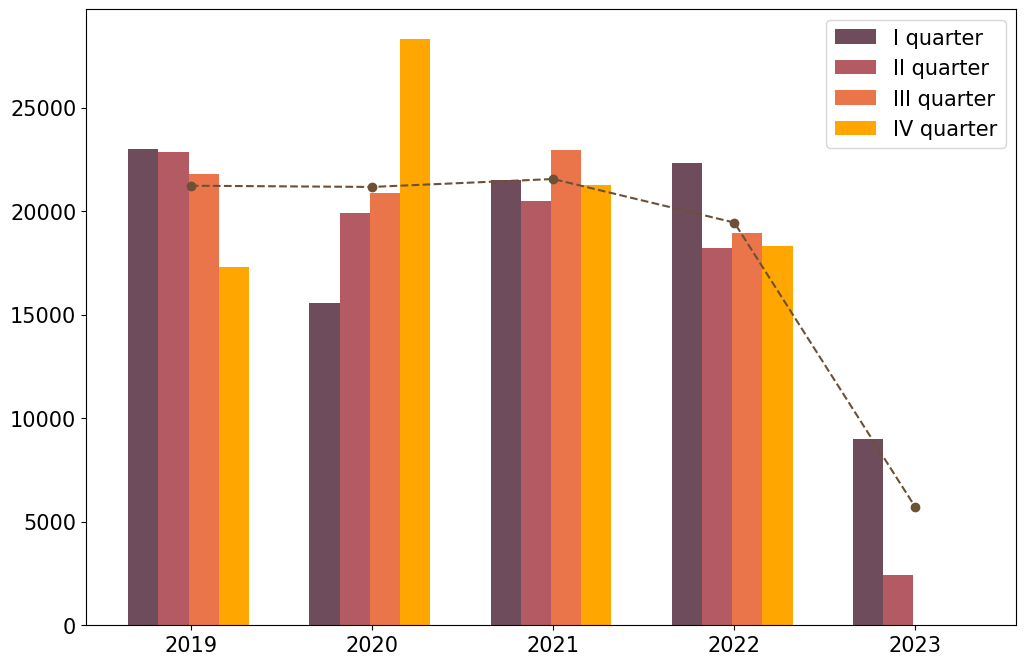}
\caption{Number of posts and comments from the "harmful" class by quarters} 
\label{timestamps}
\end{figure}

The majority of the BAN-PL dataset comprises short texts; however, it is important to note that Wykop.pl entries and comments, unlike many other social media platforms, such as Twitter, do not have a specific character limit. Consequently, user-generated content on Wykop.pl encompasses not only concise messages but also longer narrative forms, such as "copypastas" (\citet{kurcwald2015}). This factor contributes to the high standard deviation observed in the harmful class (see token statistics for specific classes in tab. \ref{tab}). In order to illustrate the evolving landscape of content moderation and its impact over time, we present a visual representation of the number of pieces of content banned by Wykop.pl moderation system in subsequent quarters from 2019 through 2023 in fig. \ref{timestamps}.

\begin{table*}[!ht]
\caption{Statistics for the BAN-PL dataset (number of tokens)}\label{tab}
\centering
\begin{tabular}{|l|r|r|r|r|r|r|}
\hline
                                                    Dataset class   & \textbf{count} & \textbf{mean} & \textbf{std} & \textbf{25\%} & \textbf{50\%} & \textbf{75\%} \\

\hline
\textbf{\begin{tabular}[c]{@{}l@{}}Harmful total (n = 345,831)\end{tabular}} & 12,064,249        & 34.88            & 98.48           & 10             & 17            & 35            \\
\textbf{\begin{tabular}[c]{@{}l@{}}Harmful flagged (n = 148,386) \end{tabular}}     & 7,392,680         & 37.44            & 118.05           & 10             & 18            & 35           \\
\textbf{\begin{tabular}[c]{@{}l@{}}Harmful predicted  (n = 197,445)\end{tabular}}   & 4,671,569     & 31.48            & 63.55            & 11             & 18            & 34            \\
\textbf{\begin{tabular}[c]{@{}l@{}}Neutral (n = 345,831) \end{tabular}}      & 14,143,837            & 40.90            & 55.87            & 13             & 24            & 47          \\
\hline
\end{tabular}
\end{table*}

\subsection{Linguistic features}
\label{section:Linguistics}

Similar to texts found on other social media platforms, the language used in Wykop.pl posts and comments is highly informal, replete with numerous grammatical and syntactical errors, rendering them challenging to process using NLP tools (\citet{tweeteval}). Simultaneously, many of these errors are intentionally crafted. Users of the platform purposely incorporate certain popular words with errors, such as "huop" instead of "chłop" [\textit{man}] or "kąkuter" instead of "komputer" [\textit{computer}]. In this case, the normalization of the text, including the correction of linguistic errors, would lead to the loss of the semantic surplus. 

Another form of semantic surplus pertains to emojis and textual emoticons formed using various Unicode and ASCII characters, alongside elements of Zalgo text. Lenny faces, a popular example, are textual emoticons utilized to convey a mischievous mood, imply sexual innuendo, or disrupt online discussions. Wykop.pl users not only employ established character combinations prevalent in internet discourse but also create their own intricately crafted ones. These combinations often elude algorithms designed to convert emoticons or emojis into textual representations. Managing emojis and emoticons during the text preprocessing stage presents a challenge when dealing with such specific data. However, certain portions of this information might be vital from a sentiment analysis standpoint. Consequently, in the published dataset, emojis and textual emoticons have been retained. Additionally, we provide a script for their removal.


In addition to spelling errors and the use of graphic characters, the language of BAN-PL, particularly in the subset of harmful content, is characterised by the extensive use of profanity, including (i) explicit, non-obfuscated profanities, (ii) profanities automatically masked during the platform's moderation process, often classified as adult content (all words, no matter the length, are transformed into the following character string "\#!\$\%@?") and (iii) profanities intentionally masked by users to evade detection and potential bans. 





Obfuscation strategies (\citet{rojas2017}), common to a number of social media platform, have been creatively developed within Wykop.pl community. Users circumvent blocking filters by obfuscating profanities with following strategies\footnote{Examples of the filters used are presented on the non-vulgar words.}:

\begin{enumerate}
\itemsep-0.1em
\item character substitution, including replacing letters with visually similar characters, especially numbers, e.g., \textit{flower -> fl0w3r}, using a string of identical symbols, usually dots or asterisks, e.g., \textit{flower -> flo...., flower --> flow**}, using random symbols, e.g., \textit{flower -> fl*\%\#\$\$r}
\item phonetic spelling (substituting letters with other letters that sound similar), e.g., \textit{duck -> duq}
\item extra character insertion or deletion, e.g., \textit{coffee -> coff\&ee; coffee -> coff*e}
\item splitting or merging, e.g.,\textit{coffee -> c o f f e e}
\end{enumerate}

While these examples represent the most common obfuscation strategies, there are numerous possible variations. Notably, some masking techniques extend beyond explicit profanities to non-vulgar words. For instance, there is a prevalence of replacing the letter "o" with a zero in certain words (e.g., "p0lka" for a Polish woman, "p0lska" for Poland, "r0sjanie" for Russians). This suggests that users may harbor negative attitudes towards these concepts, perceiving them as offensive or vulgar despite their neutral meanings. 

To facilitate dataset management, we have developed functions for unmasking obfuscated content. Firstly, we identify all profanities masked with random symbols. Next, we generate a list of potential unmasked versions based on a dictionary of profanities and the visible letters in the masked word. 
To determine the most accurate unmasked word, we compare the masked word with each proposed word from the filtered dictionary, considering symbol usage resemblance. Sorting the possible words by similarity enables us to select the closest match. The next stage involves identifying letter-to-sign and letter-to-letter exchanges. This process begins by recognizing commonly used signs in exchanges, such as \textit{@, 3, \$, q}, and others. The algorithm then attempts to create actual representations based on these popular exchanges (e.g., \textit{e->3, l->1}). Following this step, the algorithm checks whether the generated words exist in the FastText dictionary. Typically, there's only one matching word at this point, considered the correct one. However, in cases of multiple matching words, the algorithm randomly selects one since it lacks the ability to determine the best fit based on context.

Despite the inherent challenges posed by social media content, preliminary analyses utilizing corpus and computational linguistic methodology have yielded promising results in distinguishing between offensive and neutral content. Statistical analyses have revealed several features that tend to weigh towards the offensive class. Particularly noteworthy among these features are morpho-syntactical structures that support addressative forms, including the declination in the vocative case, the use of the imperative mood, as well as verbs in 2nd person singular and plural in the indicative mood, and various types of apostrophes (e.g., a noun following a 2nd person pronoun or a complex nominal phrase following a 2nd person pronoun). Furthermore, a more sophisticated task aimed at distinguishing between offensiveness and generalized hate speech has also shown potential from a strictly linguistic perspective. Some initial insights have been elaborated upon in \citet{tertium}.

\subsection{Unsupervised topic analysis}

\begin{figure}[!h]
\centering
\includegraphics[width=7.5cm]{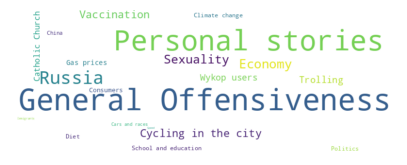}
\caption{20 topics for the anonymized BAN-PL} 
\label{small_topics}
\end{figure}

\begin{figure}[!h]
\centering
\includegraphics[width=7.5cm]{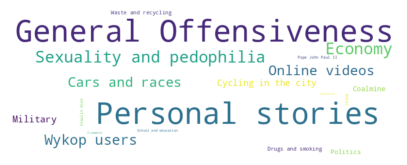}
\caption{20 topics for the entire dataset} 
\label{big_topics}
\end{figure}

In our study, we performed topic modeling on two datasets: the large set of 691,662 samples and the openly anonymized subset of 24,000 pieces. The method included BERTopic model utilizing paraphrase-multilingual-MiniLM-L12-v2 embeddings and the number of topics set to 20, with following parameters for UMAP (n\_neighbors=15, n\_components=5, min\_dist=0.0, metric='cosine', low\_memory=self.low\_memory) and HDBSCAN (min\_ster\_size=10, metric='euclidean', cluster\_selection\_method='eom', prediction\_data=True). The resulting topics were given simple labels based on keywords and their representation analysis. The topics' count was normalized to the percentage of the entire set and used for the word clouds shown in fig. \ref{small_topics} and \ref{big_topics} as the size parameter.

Both datasets prominently feature two major topics. The first, "Personal Stories," encompasses a range of life-related narratives, including observations, memories, confessions, gossip, and comments on others' life decisions, varying from neutral to sarcastic tones. The second, "General Offensiveness," aggregates a diverse set of topics under the broad category of generally offensive and hateful remarks directed at people and their lifestyles. While these two topics are consistently present in both datasets, the remaining themes exhibit more subtlety and are more susceptible to variation. However, nine specific themes—Economy, Sexuality, Urban Cycling, User Interactions on Wykop.pl, Diet, Politics, Education, China, and Cars and races—recur in both datasets. Additionally, two themes, Climate Change and Waste and Recycling, closely align, weaving similar threads from slightly different perspectives.


\subsection{Preliminary experiments}

To evaluate the effectiveness of the proposed dataset for binary classification task, we conducted a series of preliminary experiments using four distinct BERT-based models for the Polish language, with two of them being trained for identifying specific types of offensiveness, namely cyberbullying and hate speech. The RoBERTa base v2 (RoBERTa), TrelBERT, Polbert-CB,\footnote{\url{https://huggingface.co/ptaszynski/bert-base-polish-cyberbullying}} and Polish-Hate-Speech-Detection-Herbert-Large (HerBERT-HS\footnote{\url{https://huggingface.co/dkleczek/Polish-Hate-Speech-Detection-Herbert-Large}}) model were implemented using the PyTorch framework, AdamW as the optimizer and learning rate = 5e-5. 

All models were fine-tuned on a split of 80\% of the open BAN-PL dataset for training, 10\% for validation, and tested on the remaining 10\%, giving the support count of 2400 samples (with 1191 for neutral and 1209 for the offensive class). The only pre-processing of the data included the removal of new-line coding '\texttt{\textbackslash n}'
and inserting additional spaces in the apostrophes between the addressed user nickname token '{USERNAME}:' and the following utterance, for leaving it merged into a long string could bring noise to the model. 

Early stopping was employed, monitoring the validation loss with a patience of 5 epochs. Tab. \ref{tab} summarizes the results obtained from the preliminary experiments based on recall, precision, and F1 scores for the offensive class. As expected, fine-tuning a base model proved more efficient than fine-tuning an already fine-tuned model, which despite a similar task was initially trained on a different dataset involving cyberbullying or hate speech. 

\begin{table}[ht]
\caption{The Results of the preliminary experiments on the open BAN-PL dataset}\label{tab}
\centering
\begin{tabular}{|l|r|r|r|r|r|r|}
\hline
                                                    Model   & \textbf{Recall} & \textbf{Precision} & \textbf{F1} \\             
\hline
\textbf{\begin{tabular}[c]{@{}l@{}}RoBERTa base v2\end{tabular}} & 0.84        & 0.83            & 0.83           \\
\textbf{\begin{tabular}[c]{@{}l@{}}TrelBERT \end{tabular}}     & 0.76         & 0.82            & 0.79  \\
\textbf{\begin{tabular}[c]{@{}l@{}}Polbert-CB\end{tabular}}   & 0.83   & 0.78            & 0.81                 \\
\textbf{\begin{tabular}[c]{@{}l@{}}HerBERT-HS\end{tabular}}      & 0.79            & 0.80            & 0.79           \\
\hline
\end{tabular}
\end{table}

\section{Discussion}
\label{discussion}

Based on the literature review and the evaluation of the leading datasets currently assigned to the task of offensive language detection for Polish and English, one can identify the following biases: (i) post-moderation source bias, (ii) pre-selection bias, (iii) annotation bias. The first two notions refer directly to data collection process, in which harmful data is usually obtained from publicly available social media content (most notably Twitter) through APIs based on pre-selection of users, key words or hashtags (e.g., ~\citet{poleval_ptaszynski, Davidson}). As a consequence, obtained data might have already undergone some initial moderation resulting in limited number of explicit hateful posts. Furthermore, these approaches still undermine the possibility of creating a genuinely representative corpus of abusive content by collecting relatively homogeneous data centered around specific topics or targets of hate (\citet{Ludwig-2022}). User distribution is also of crucial value. It has been successfully demonstrated that user distribution bias, resulting from obtaining huge amounts of data from single users, is also present in often-cited English datasets (\citet{Arango-2022}). Annotation bias covers a range of challenges resulting from varying annotation guidelines, which are rooted in the lack of consensus on definitions of harmful content and subjective notions of what constitutes hate speech (\citet{Ross-2016, waseem2016hateful}), different levels of demographic diversity of annotators, and the need of reliable inter-annotator agreement. In this context, the problem of personal bias has been highlighted (\citet{Sap-2019, Sap-2021}, see also \cite{Garg-2022}).

When evaluating BAN-PL dataset from the perspective of the above mentioned biases, it is evident that the first two challenges posed by data collection are successfully mitigated. The dataset comprises a huge amount of harmful content, which was possible due to obtaining data directly from the moderation team. Another advantage of this approach is that the harmful content has not been pre-selected, but comes from a relatively long time span and covers multiple topics, not limited to the ones most commonly considered hate-related. Additionally, the harmful content was generated by a large number of unique users, therefore it is free of user distribution bias. Annotation bias cannot be easily comparable betweeen BAN-PL and other annotated datasets, since moderation relied on the website's internal policy and was conducted by professionals. Thus, we do not have control over annotation process or insight into moderators' demographic information due to required anonymity and cannot evaluate possible inherent personal biases. However, voting system employed within the moderation process facilitates the mitigation of personal biases. 

The moderation process itself, as made evident in the task of re-annotating KLEJ harmful content by Wykop.pl moderators (see \ref{subsection:moderation}), can hinder the generalization of the model trained on BAN-PL. The very specific linguistic nature of the content along with users' demographics which differs from other popular social media platforms can also have a negative impact on generalization, as established in regard to other datasets (\citet{waseem2018bridging}).  

A major advantage of the BAN-PL dataset is the coverage of multiple topics and hate-related concepts, including personal attacks, generalized hate speech, and toxicity, as well as a wide range of targets of attacks, such as public figures, women, and other protected groups. However, for the purpose of analyzing specific aspects of hateful language, one would need to distinguish further subsets, which require additional manual annotation. Such umbrella terms, as offensive language or cyberbullying, cover a wider range of more nuanced phenomena, which can be considered overlapping. Therefore, binary approach is often employed, however it may also be considered too simplistic (\citet{Sang-2022}).
As opposed to the most commonly utilized datasets of offensive language (\citet{indatawetrust}), BAN-PL provides equal classes of harmful and neutral content. Data collection aimed at obtaining content which would prove successfully comparable in terms of topics covered, however the data comes from slightly different time spans and may result in overrepresentation of certain trending topics in the harmful class.  

\section{Conclusion and future work}

This paper presents several notable contributions: 
\begin{enumerate}
\itemsep-0.07em

    \item introduction of a new open dataset tailored for offensive language detection in the Polish language. Unlike existing datasets, this collection comprises texts originally posted online, subsequently flagged by users, and moderated by professionals for banning. Consequently, it represents data that was previously inaccessible, now reconstructed to form an intentional dataset.
    \item the analysis of significant linguistic features of the content posted on Wykop.pl, which are relevant when working with the dataset;
    \item the identification of biases that the dataset effectively avoids and those that still persist;
    \item a discussion of potential strategies for addressing these biases in datasets based on content removed during the moderation process;
    \item the publication of the anonymized open available subset of the dataset along with preprocessing scripts that can be readily applied in real-life scenarios.
\end{enumerate}

Additionally, this paper provides valuable insights into anonymization guidelines and offers an evaluation of the moderation framework.

However, to address further challenges and improve the analysis of the offensive language, we aim to continue developing preprocessing functions for text normalization and profanity unmasking. Moreover, we aim to advance scholarly understanding of online offensive behavior, with a particular emphasis on studying hate speech directed at groups based on protected characteristics. We aim at releasing a subset of hate speech, which would require manual annotation. This task will facilitate providing a dictionary of Polish expressions related to hate speech and offensive language that can be further applied to a plethora of NLP tasks. The manually annotated and evaluated subset of 24,000 samples will be used to fine-tune the NER model. Ultimately, the anonymization of the whole dataset will be done in an iterative human-in-the-loop manner, using a refined automatic pipeline and involving decreasing manual correction. 

\section{Data availability}

Owing to the comprehensive anonymization procedure (see \ref{section:anonymization}), the complete dataset will be incrementally released. The initial segment of publication comprises 24,000 samples, evenly distributed between harmful and neutral classes, and has already made available on our GitHub account along with preprocessing scripts. All forthcoming releases, including the model and anonymizer, will also be made available on our GitHub account to ensure transparency and facilitate the reproducibility of our research.\footnote{\url{https://github.com/ZILiAT-NASK/BAN-PL}}

\section{Ethical considerations}

The authors of this paper present a novel dataset focusing on offensive language, cyberbullying, and hate speech in the Polish language. The data was sourced from the popular social news platform Wykop.pl, in collaboration with Wykop.pl moderators. The collected samples were identified as violations of the website's internal policy by the community of users and moderators. The authors argue that this dataset offers distinct advantages compared to resources obtained through public APIs and manual annotation processes. In compliance with the Association for Internet Researchers Ethical Guidelines,\footnote{\url{https://aoir.org/reports/ethics3.pdf}} the authors conducted a thorough anonymization procedure. This process ensured that no screen names of authors and sensitive personal data, such as phone numbers and identification numbers, were publicly available. Additionally, the targets of insults, including public figures like politicians or celebrities, were also anonymized to restrict the dissemination of hateful content.

\nocite{*}
\section{Bibliographical References}\label{sec:reference}

\bibliographystyle{lrec-coling2024-natbib}
\bibliography{BAN}


\end{document}